\documentclass[lettersize,journal]{IEEEtran}
\usepackage{amsmath,amsfonts}
\usepackage{algorithmic}
\usepackage{array}
\usepackage[caption=false,font=normalsize,labelfont=sf,textfont=sf]{subfig}
\usepackage{textcomp}
\usepackage{stfloats}
\usepackage{url}
\usepackage{verbatim}
\usepackage{graphicx}
\usepackage{multirow}
\usepackage{multicol}
\def\BibTeX{{\rm B\kern-.05em{\sc i\kern-.025em b}\kern-.08em
    T\kern-.1667em\lower.7ex\hbox{E}\kern-.125emX}}
\usepackage{balance}
\usepackage{booktabs}
\PassOptionsToPackage{table, dvipsnames}{xcolor}
\usepackage[table]{xcolor}
\usepackage[dvipsnames]{xcolor}
\usepackage{colortbl}

\usepackage{float}
\usepackage[pagebackref,breaklinks,colorlinks]{hyperref}
\begin{document}
\title{Panacea+: Panoramic and Controllable Video Generation for
Autonomous Driving}
\author{Yuqing Wen, Yucheng Zhao, Yingfei Liu, Binyuan Huang, Fan Jia, Yanhui Wang, Chi Zhang, Tiancai Wang, Xiaoyan Sun, Xiangyu Zhang \\
\textit{Project Page}:~\href{https://panacea-ad.github.io/}{https://panacea-ad.github.io/} 
\thanks{The work was supported by National Science and Technology Major Project of China
(2023ZD0121300).\\
Yuqing Wen, Yanhui Wang and Xiaoyan Sun are with the School of
Information Science and Technology, University of Science and Technology of China, Hefei, Anhui, 230026, China (email: wenyuqing@mail.ustc.edu.cn; wangyanhui@mail.ustc.edu.cn; sunxiaoyan@ustc.edu.cn) \\
Yucheng Zhao, Yingfei Li, Fan Jia, Tiancai Wang and Xiangyu Zhang are with MEGVII Technology, Beijing, 100191, China (email: zhaoyucheng@megvii.com; liuyingfei@megvii.com; jiafan@megvii.com; wangtiancai@megvii.com; zhangxiangyu@megvii.com)\\
Binyuan Huang is with the School of Remote Sensing and
Information Engineering, Wuhan University, Wuhan, Hubei, 430072, China (email: byuan@whu.edu.cn )\\
Chi Zhang is with Mach Drive, Wuhu, Anhui, 241004, China (email: chi.zhang@mach-drive.com)

}}

\maketitle

\begin{abstract}

The field of autonomous driving increasingly demands high-quality annotated video training data. In this paper, we propose Panacea+, a powerful and universally applicable framework for generating video data in driving scenes. Built upon the foundation of our previous work, Panacea, Panacea+ adopts a multi-view appearance noise prior mechanism and a super-resolution module for enhanced consistency and increased resolution. Extensive experiments show that the generated video samples from Panacea+ greatly benefit a wide range of tasks on different datasets, including 3D object tracking, 3D object detection, and lane detection tasks on the nuScenes and Argoverse 2 dataset. These results strongly prove Panacea+ to be a valuable data generation framework for autonomous driving.

\end{abstract}

\renewcommand{\thefootnote}{\fnsymbol{footnote}}
\footnotetext[1]{This work has been submitted to the IEEE for possible publication. Copyright may be transferred without notice, after which this version may no longer be accessible.}


\section{Introduction}

\IEEEPARstart{I}{n} the field of autonomous driving, Bird's Eye View (BEV) perception methods have attracted significant research interest in recent years, demonstrating notable potential across key perception tasks, including 3D object detection~\cite{DBLP:journals/corr/abs-2112-11790, DBLP:journals/corr/abs-2303-11926, DBLP:conf/eccv/LiWLXSLQD22}, map segmentation~\cite{liu2023petrv2, jiang2023polarformer}, multi-object tracking~\cite{DBLP:conf/cvpr/PangLT0ZW23, DBLP:journals/corr/abs-2303-11926}, and 3D lane detection~\cite{chen2022persformer, huang2023anchor3dlane}. Typically, video-based BEV perception methods, like StreamPETR~\cite{DBLP:journals/corr/abs-2303-11926}, usually exhibit better performance than image-based methods in these perception tasks, especially in tracking, which requires understanding the temporal coherence of objects in video sequences. To achieve superior performance, these video-based methods generally demand large-scale and high-quality video data for training.
However, in practical scenarios, acquiring such data poses significant challenges. Traditional manual data collection processes not only incur extremely high costs but also suffer from a lack of data diversity due to safety concerns. For instance, collecting data in extreme weather conditions is virtually impossible.

Given the aforementioned issues, many studies focus on synthesizing viable samples to augment existing datasets for autonomous driving. 
Some efforts concentrate on generating image data conditioned with layouts, as seen in works like~\cite{DBLP:journals/corr/abs-2308-01661, DBLP:journals/corr/abs-2301-04634, DBLP:conf/cvpr/ZhengZLQSL23, DBLP:journals/corr/abs-2302-08908}. Other methods address the more complex task of synthesizing video data to benefit the more advanced video-based perception methods. For instance, MagicDrive~\cite{DBLP:journals/corr/abs-2310-02601}, DriveDreamer~\cite{DBLP:journals/corr/abs-2309-09777}, and Wovogen~\cite{DBLP:journals/corr/abs-2312-02934} generate video samples based on BEV layout sequences. Our previous work, Panacea~\cite{DBLP:journals/corr/abs-2311-16813}, a diffusion-based~\cite{DBLP:conf/cvpr/RombachBLEO22} video generation model, also targets this challenging task. It produces multi-view videos of high fidelity in driving scenarios aligned with BEV layout sequences and textual control signals, effectively benefiting perception models on object detection task.

However, Panacea still has room for improvement before becoming more practical. In applications, synthetic training data with high temporal consistency is essential, especially for tasks like tracking, which rely on temporal modeling. In this respect, Panacea still leaves potential for enhancement. Moreover, it has been demonstrated that training perception models with high-resolution data can significantly enhance the performance~\cite{DBLP:journals/corr/abs-2303-11926}. Nevertheless, Panacea currently generates samples at a relatively low resolution of 256 $\times$ 512. Using such synthetic data to augment perception models imposes a limit on the potential performance upper bond.

\begin{figure}
  \centering
   \includegraphics[width=1.0\linewidth]{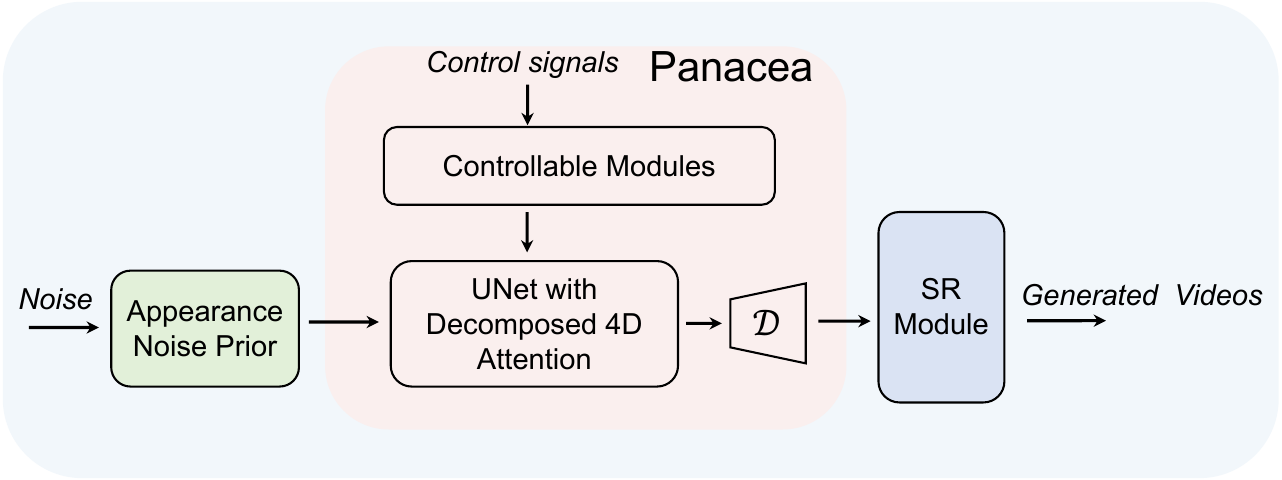}
   \hfill
   \caption{Framework outline of Panacea+. Panancea+ is built upon Panacea while introducing an appearance noise prior and a super-resolution module. Here "D" denotes the VAE decoder.}
   \label{frameplus}
   \vspace{-0.5cm}
\end{figure}

\begin{figure*}[h]
  \centering
   \includegraphics[width=\linewidth]{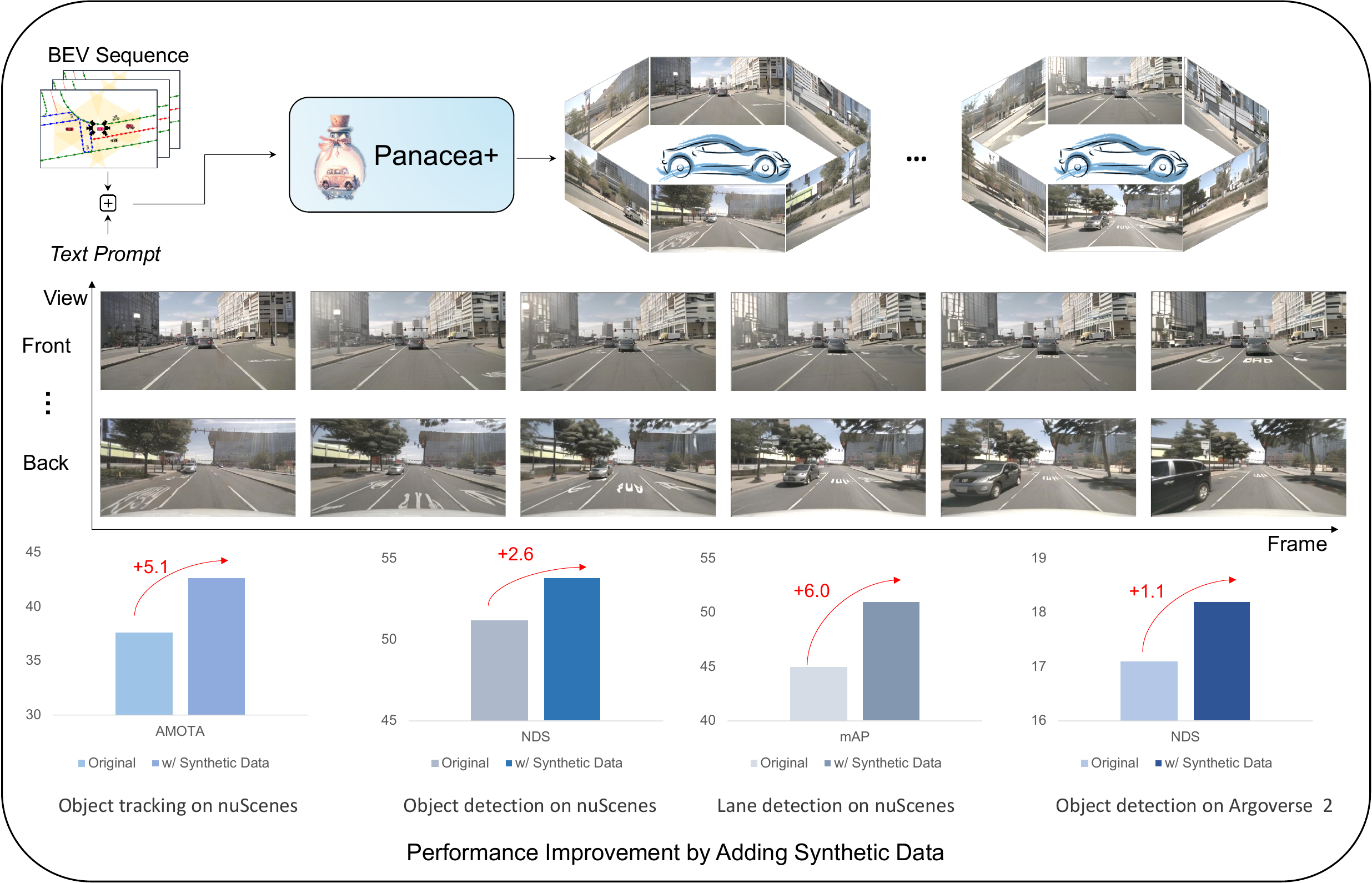}
   \hfill
    \caption{\textbf{Visualizations of the Panacea+'s capability.} Panacea+ is able to generate high-quality and high resolution multi-view videos with BEV layout control and textual control. The generated samples can greatly benefit many tasks on different datasets in autonomous driving including object tracking, object detection, and lane detection, verifying its power and versatility for autonomous driving.
   }
   \label{fig:train}
\end{figure*}

In this paper, we introduce Panacea+, an enhanced version of Panacea. Panacea+ is designed to transform Panacea into a more powerful and practical data generation framework tailored for autonomous driving applications. Built upon the structure of Panacea, Panacea+ integrates multi-view appearance noise prior to enhance consistency and a super-resolution module to enable higher resolution synthesis, as illustrated in Fig.~\ref{frameplus}. Specifically, Panacea+ includes a two-stage video generation model that first synthesizes multi-view images and then constructs multi-view videos from these images. To achieve effective multi-view and temporal modeling, Panacea+ employs a decomposed 4D attention mechanism. Multi-view appearance noise priors are incorporated during training and inference to enhance consistency. For controllable generation, Panacea+ integrates BEV layout as control signals through a ControlNet~\cite{DBLP:journals/corr/abs-2302-05543}. Additionally, a super-resolution module is cascaded after the generation modules to increase the resolution of generated samples. These designs allow Panacea+ to synthesis high-quality, well-annotated multi-view video samples.

Furthermore, we conduct extensive experiments across multiple tasks and datasets for comprehensive validation. Panacea+ is applied to both the nuScenes~\cite{DBLP:conf/cvpr/CaesarBLVLXKPBB20} and Argoverse 2~\cite{DBLP:journals/corr/abs-2301-00493} dataset and evaluated on tasks including 3D object tracking, 3D object detection, and lane detection. These experimental results show that the video samples from Panacea+ significantly benefit the training of perception models, achieving state-of-the-art (SOTA) performance among current generation approaches. The superior performance validates Panacea+'s significant value to autonomous driving. In summary, this paper makes the following contributions:

\begin{itemize}

\item We introduce an improved version of Panacea, named Panacea+, a data generation framework including a two-stage video generation module with decomposed 4D attention and multi-view appearance noise prior for consistent generation and a ControlNet for controllability. A super-resolution module is further adopted for high-resolution synthesising. This cost-efficient framework of generating low resolution samples first and then expanding them to higher resolution is effective for enhancing BEV perception models.

\item We conduct comprehensive evaluation to validate of the effectiveness of Panacea+. Extensive experiments are conducted under a broad setting that encompasses multiple datasets and tasks. Among the current approaches, Panacea+ elaborates perception models, achieving state-of-the-art performance with an AMOTA of 42.7 in tracking and an NDS of 53.8 in detection. These superior results demonstrate that Panacea+ is a powerful and universally applicable generation framework for autonomous driving.

\end{itemize}

\section{Related works}
\subsection{Diffusion-based Generative Models}
Diffusion models (DMs) have achieved remarkable advancements in image generation~\cite{DBLP:conf/icml/NicholDRSMMSC22, DBLP:journals/corr/abs-2204-06125,DBLP:conf/nips/SahariaCSLWDGLA22,DBLP:journals/corr/abs-2307-01952, DBLP:conf/nips/DhariwalN21}. In particular, Stable Diffusion (SD)~\cite{DBLP:conf/cvpr/RombachBLEO22} employs DMs within the latent space of autoencoders, striking a balance between computational efficiency and high image quality, marking a milestone in image generation. In addition to text conditioning, the field is advancing with the introduction of supplementary control signals~\cite{DBLP:journals/corr/abs-2302-05543, DBLP:journals/corr/abs-2302-08453, 10547051}. A notable example is ControlNet~\cite{DBLP:journals/corr/abs-2302-05543}, which incorporates a trainable copy of the Stable Diffusion (SD) encoder to integrate control signals. Additionally, some studies focus on generating multi-view images. For instance, MVDffusion~\cite{DBLP:journals/corr/abs-2307-01097} processes perspective images in parallel using a pretrained diffusion model.

Beyond image generation, video generative diffusion models \cite{DBLP:journals/corr/abs-2210-02303,DBLP:journals/corr/abs-2212-11565, DBLP:journals/corr/abs-2211-11018, DBLP:journals/corr/abs-2310-10769, DBLP:journals/corr/abs-2305-10474, DBLP:journals/corr/abs-2308-06571, DBLP:journals/corr/abs-2309-00398} are gaining significant attention and have made remarkable advancements recently. Earlier works primarily concentrated on generating short video clips~\cite{DBLP:conf/cvpr/BlattmannRLD0FK23,DBLP:journals/corr/abs-2211-11018,DBLP:conf/iclr/SingerPH00ZHYAG23}. For example, MagicVideo~\cite{DBLP:journals/corr/abs-2211-11018} uses frame-wise adaptors and a causal temporal attention module for text-to-video generation. The Video Latent Diffusion Model (VLDM)~\cite{DBLP:conf/cvpr/BlattmannRLD0FK23} integrates temporal layers into a 2D diffusion model to produce temporally aligned videos. Make-A-Video \cite{DBLP:conf/iclr/SingerPH00ZHYAG23} extends a diffusion-based text-to-image model without requiring text-video pairs. Imagen Video~\cite{DBLP:journals/corr/abs-2210-02303} employs a chain of video diffusion models to generate videos from text inputs. Stable Video Diffusion~\cite{DBLP:journals/corr/abs-2311-15127}, trained on large-scale datasets of short video clips, is capable of synthesizing high-quality short videos. Afterwards, the emergence of Sora~\cite{sora}, demonstrating the potential of diffusion models for high-quality long video generation.

Our method also pertains to video generation, but unlike previous approaches, we focus on a more complex and specific scenario, creating controllable multi-view videos in driving contexts.

\subsection{Generation for Autonomous Driving}

The development of Bird's-Eye-View (BEV) representation \cite{DBLP:conf/eccv/LiWLXSLQD22, liu2022petr,DBLP:conf/aaai/LiGYYWSSL23,DBLP:journals/corr/abs-2303-11926, DBLP:conf/aaai/JiangLLWJWHZ24F} in multi-view perception has become a pivotal area of research in autonomous driving. This advancement plays a crucial role in improving downstream tasks such as multi-object tracking \cite{DBLP:conf/cvpr/ZhangCWWZ22, DBLP:conf/cvpr/PangLT0ZW23}, motion prediction \cite{DBLP:conf/cvpr/PangLT0ZW23}, and planning \cite{DBLP:conf/cvpr/HuYCLSZCDLWLJLD23,DBLP:journals/corr/abs-2303-12077}. 
These tasks typically favor temporal information, especially tracking that relys heavily on the consistency of data over time. Recently, video-based BEV perception methods \cite{DBLP:conf/eccv/LiWLXSLQD22,DBLP:conf/aaai/LiGYYWSSL23,DBLP:conf/iclr/ParkXYKKTZ23,DBLP:journals/corr/abs-2303-11926} have become dominant. BEVFormer \cite{DBLP:conf/eccv/LiWLXSLQD22} was a pioneer in integrating temporal modeling mechanisms, significantly improving upon single-frame methods like DETR3D \cite{DBLP:conf/corl/WangGZWZ021} and PETR \cite{liu2022petr}. Following this, BEVDepth \cite{DBLP:conf/aaai/LiGYYWSSL23}, SOLOFusion \cite{DBLP:conf/iclr/ParkXYKKTZ23}, and StreamPETR \cite{DBLP:journals/corr/abs-2303-11926} have further enhanced temporal modeling approaches, leading to superior performance. In addition, various experimental results indicating that these perception models typically favor high resolution data for training.

As BEV perception methods heavily depend on high quality paired data with accurate BEV ground truth layouts, numerous studies are investigating paired data generation to support training. Previously, some generative efforts in autonomous driving have utilized BEV layouts to augment image data with synthetic single or multi-view images \cite{DBLP:journals/corr/abs-2308-01661, DBLP:journals/corr/abs-2301-04634}, proving beneficial for single-frame perception methods. For example, BEVGen \cite{DBLP:journals/corr/abs-2301-04634} specializes in generating multi-view street images based on BEV layouts, while BEVControl \cite{DBLP:journals/corr/abs-2308-01661} proposes a two-stage generative pipeline for creating image foregrounds and backgrounds from BEV layouts. Further studies are exploring the generation of paired video data, which is crucial for more advanced video-based BEV perception methods.
The Video Latent Diffusion Model \cite{DBLP:conf/cvpr/BlattmannRLD0FK23} attempts to generate driving videos but its scope is limited to single-view and it falls short in effectively bolstering video-based perception models. Magicdrive~\cite{DBLP:journals/corr/abs-2310-02601} synthesizes multi-view videos but lacks exploration into the effectiveness of temporal-based BEV perception methods. DriveDreamer~\cite{DBLP:journals/corr/abs-2309-09777} and DriveDreamer-2~\cite{DBLP:journals/corr/abs-2403-06845} synthesize multi-view video data based on diverse control signals. Panacea~\cite{DBLP:journals/corr/abs-2311-16813} explores generating multi-view videos paired with BEV layout sequences and validate on temporal-based BEV methods. SubjectDrive~\cite{DBLP:journals/corr/abs-2403-19438} builds upon Panacea by introducing subject control and explores the impact of synthetic data scaling on perception tasks. 

In this paper, the proposed Panacea+ aims to provide a versatile and powerful generation framework that can generate high-quality and high resolution videos for autonomous driving.

\begin{figure*}[t]
  \centering
   \includegraphics[width=\linewidth]{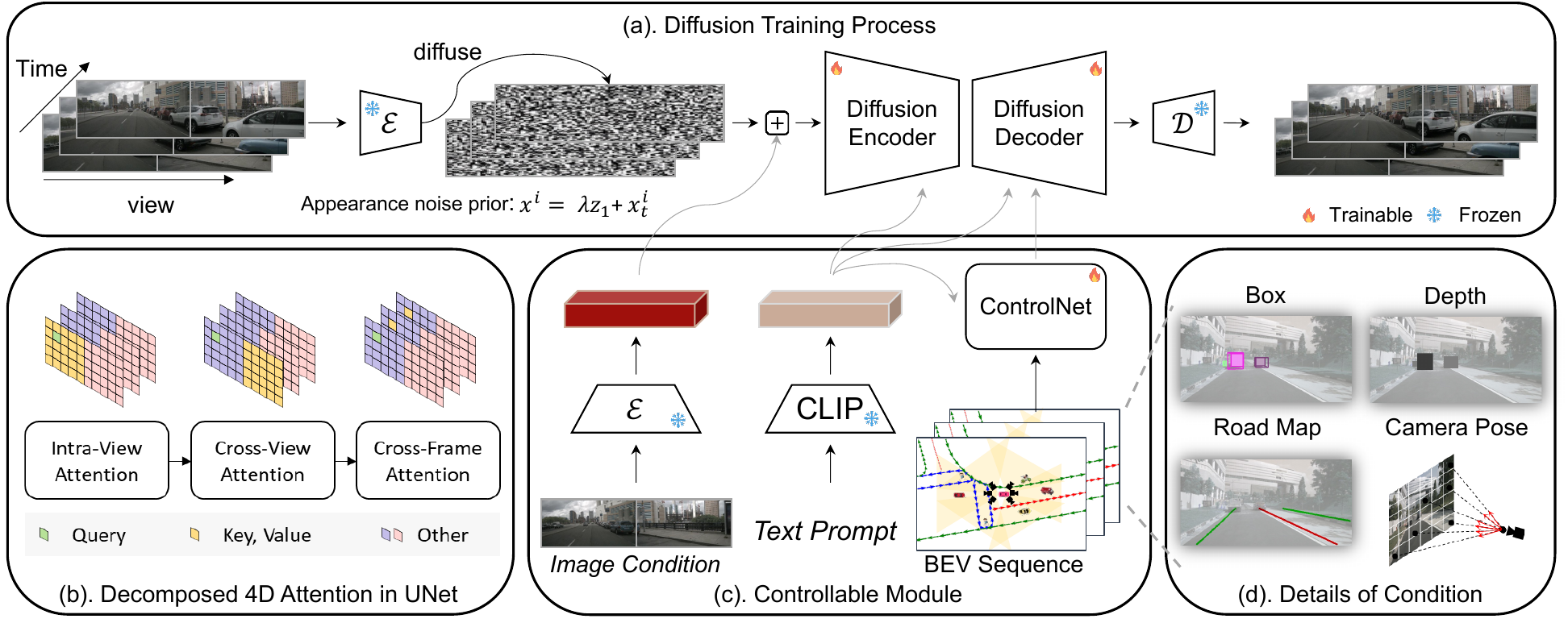}
   \hfill
    \caption{\textbf{Overview of Panacea+.} (a). The diffusion training process of Panacea+, enabled by a diffusion encoder and decoder with the decomposed 4D attention module. An appearance noise prior is applied for enhancing temporal consistency. 
    (b). The decomposed 4D attention module comprises three components: intra-view attention for spatial processing within individual views, cross-view attention to engage with adjacent views, and cross-frame attention for temporal processing. 
    (c). Controllable module for the integration of diverse signals. The image conditions are derived from a frozen VAE encoder and combined with diffused noises. The text prompts are processed through a frozen CLIP encoder, while BEV sequences are handled via ControlNet. 
    (d). The details of BEV layout sequences, including projected bounding boxes, object depths, road maps and camera poses.
   }
   \label{fig:train}
\end{figure*}

\section{Method}

In this section, we present the detailed methodology of Panacea+ for generating controllable multi-view videos for driving scenes. Panacea+ builds on the foundational framework of Panacea. However, we still provide a detailed description of our method to ensure comprehensiveness. Sec. \ref{preliminary} provides a brief description of the latent diffusion models that form the foundation of our approach. Following this, Sec. \ref{framework} delves into the detail of our method that enables the generation of high-quality multi-view videos in a feasible and efficient manner. Finally, Sec. \ref{control} elaborates on the controlling modules integral to Panacea+, which is the central feature that enabling the generation of annotated training samples for the advancement of autonomous driving systems.

\subsection{Preliminary: Latent Diffusion Models} \label{preliminary}

\noindent\textbf{Diffusion models (DMs)}~\cite{DBLP:conf/nips/HoJA20, DBLP:conf/iclr/SongME21} learn to approximate a data distribution $p(x)$ via iteratively denoising a normally distributed noise $\epsilon$. Specifically, DMs first construct the diffused inputs $x_t$ through a fixed forward diffusion process in Eq.~\ref{eq:add noise}. Here $\alpha_t$ and $\sigma_t$ represent the given noise schedule, and $t$ indicates the diffusion time step. Then, a denoiser model $\epsilon_\theta$ is trained to estimate the added noise $\epsilon$ from the diffused inputs $x_t$. This is achieved by minimizing the mean-square error, as detailed in Eq.~\ref{eq:objective}. Once trained, DMs are able to synthesize a new data $x_0$ from random noise $x_T\sim \mathcal{N}(\mathbf{0}, \boldsymbol{I}) $ by sampling $x_t$ iteratively, as formulated in Eq.~\ref{eq:distribution}. Here $\mu_{\theta}$ and $\Sigma_{\theta}$ are determined through the denoiser model $\epsilon_\theta$ \cite{DBLP:conf/nips/HoJA20}.
\begin{equation}
\label{eq:add noise}
x_{t}=\alpha_{t} x+\sigma_{t} \epsilon,  \epsilon \sim \mathcal{N}(\mathbf{0}, \boldsymbol{I}), x \sim p(x)
\end{equation}
\begin{equation}
\label{eq:objective}
\min_{\theta}\mathbb{E}_{t,x,\epsilon}||\epsilon-\epsilon_{\theta}(\mathbf{x}_{t}, t)||_{2}^{2}
\end{equation}
\begin{equation}
\label{eq:distribution}
    p_{\theta}\left(x_{t-1} \mid x_{t}\right)=\mathcal{N}\left(x_{t-1} ; \mu_{\theta}\left(x_{t}, t\right), \Sigma_{\theta}\left(x_{t}, t\right)\right)
\end{equation}

\noindent\textbf{Latent diffusion models (LDMs) }~\cite{DBLP:conf/cvpr/RombachBLEO22} are a variant of diffusion models that operate within the latent representation space rather than the pixel space, effectively simplifying the challenge of handling high-dimensional data. This is achieved by transforming pixel-space image into more compact latent representations via a perceptual compression model. Specifically, for an image \( x \), this model employs an encoder \( \mathcal{E} \) to map \( x \) into the latent space \( z = \mathcal{E}(x) \). This latent code \( z \) can be subsequently reconstructed back to the original image \( x \) through a decoder \( \mathcal{D} \) as \( x = \mathcal{D}(z) \). The training and inference processes of LDMs closely mirror those of traditional DMs, as delineated in Eq. \ref{eq:add noise}-\ref{eq:distribution}, except for the substitution of \( x \) with the latent code \( z \).

\subsection{Generating High-Quality Multi-View Videos}\label{framework}

In this section we describe how Panacea+ generate high-quality multi-view videos based on a pre-trained image LDM~\cite{DBLP:conf/cvpr/RombachBLEO22}. Our model utilizes a multi-view video dataset $p_{data}$ for training. Each video sequence, encompasses \( T \) frames, indicating the sequence length, \( V \) different views, and dimensions \( H \) and \( W \) for height and width, respectively.

Panacea+ is built on Stable Diffusion (SD)~\cite{DBLP:conf/cvpr/RombachBLEO22}, a strong pre-trained latent diffusion model for image synthesis. While the SD model excels in image generation, its direct application falls short in producing consistent multi-view videos due to the lack of constraints between different views and frames in the sequence. To simultaneously model temporal and spatial consistency, the decomposed 4D attention-based UNet is adopted to concurrently generate the entire multi-view video sequence. The joint diffused input \( z \) is structured with dimensions \( H \times (W \times V) \times T \times C \), where $C$ represents the latent dimension. This multi-view video sequence is constructed by concatenating the frames across their width, which aligns with their inherent panoramic nature. Fig.~\ref{fig:train} (a) illustrates the overall training framework of Panacea+. Additionally, a multi-view appearance noise prior is adopted for better consistensy. Beyond the 4D attention-based UNet, a two-stage generation pipeline is used to largely boost the generation quality. Furthermore, in order to explore the effectiveness of high-resolution training for perception tasks, we integrate a super-resolution module into the overall framework. In this way, our framework is capable of synthesizing high-resolution, high-quality video data, thereby providing robust assistance for perception tasks.

\subsubsection{Decomposed 4D Attention}

The decomposed 4D attention is designed to simultaneously model view and temporal consistency while maintaining computational efficiency. It breaks down the original memory-intensive 4D joint attention into a more efficient architecture, inspired by recent explorations in video representation learning~\cite{zhao2023td, DBLP:conf/icml/BertasiusWT21,DBLP:conf/iccv/Arnab0H0LS21,DBLP:conf/eccv/LinGZGMWDQL22, DBLP:journals/tvt/ShengFXS20}. The decomposed 4D attention selectively retains the most critical components: the attention between adjacent views and the attention among spatially aligned temporal patches. This leads to two attention modules—cross-view attention and cross-frame attention—alongside the existing intra-view spatial attention.

Fig.~\ref{fig:train} (b) details the decomposed 4D attention mechanism. The intra-view attention retains the design of the original spatial self-attention in the Stable Diffusion (SD) model, as formulated in Eq.~\ref{eq:intra}. To enhance cross-view consistency, cross-view attention is introduced. Our observations indicate that the correlation between adjacent views is paramount, while the correlation among non-adjacent views is comparatively less significant and can be disregarded. This cross-view attention is formulated in Eq.~\ref{eq:inter}. The cross-frame attention, mirroring the design of VLDM~\cite{DBLP:conf/cvpr/BlattmannRLD0FK23}, focuses on spatially aligned temporal patches. This component is crucial in endowing the model with temporal awareness, a key factor in generating temporally coherent videos.
\begin{equation}
\label{eq:intra}
\mathrm{Att}_{iv}\left(Q, K, V\right)=\operatorname{softmax}\left(\frac{Q_t^{v}\left(K_t^{v}\right)^{T}}{\sqrt{c}}\right) V_t^{v}
\end{equation}

\begin{equation}
\begin{aligned}
\label{eq:inter}
\mathrm {Att}_{cv}\left(Q, K, V\right) &= \operatorname{softmax}\left(\frac{Q_t^{v}\left([K_t^{v-1},K_t^{v+1}]\right)^{T}}{\sqrt{c}}\right) \\
&\quad \cdot [V_t^{v-1},V_t^{v+1}]
\end{aligned}
\end{equation}

Here $Q_t^{v},K_t^{v},V_t^{v}$ represent the queries, keys, and values within frame $t$ and view $v$, respectively.

\subsubsection{Two-Stage Pipeline}
\label{two stage}
To enhance the generation quality, we further adopt a two-stage training and inference pipeline. By bypassing the temporal-aware modules, our model could also operate as a multi-view image generator, which enables a unified architecture for two-stage video generation.

During training, we first train a separate set of weights dedicated to multi-view image generation. Then, as illustrated in Fig.~\ref{fig:train}, we train the second stage video generation weights, by concatenating a conditioned image alongside the diffused input. This conditioned image is integrated only with the first frame, while subsequent frames employ zero padding. Notably, in this second stage training, we employ ground truth images instead of the generated ones as condition. This approach equips our training process with an efficiency comparable to that of a single-stage video generation scheme.

\begin{figure}
  \centering
   \includegraphics[width=1.0\linewidth]{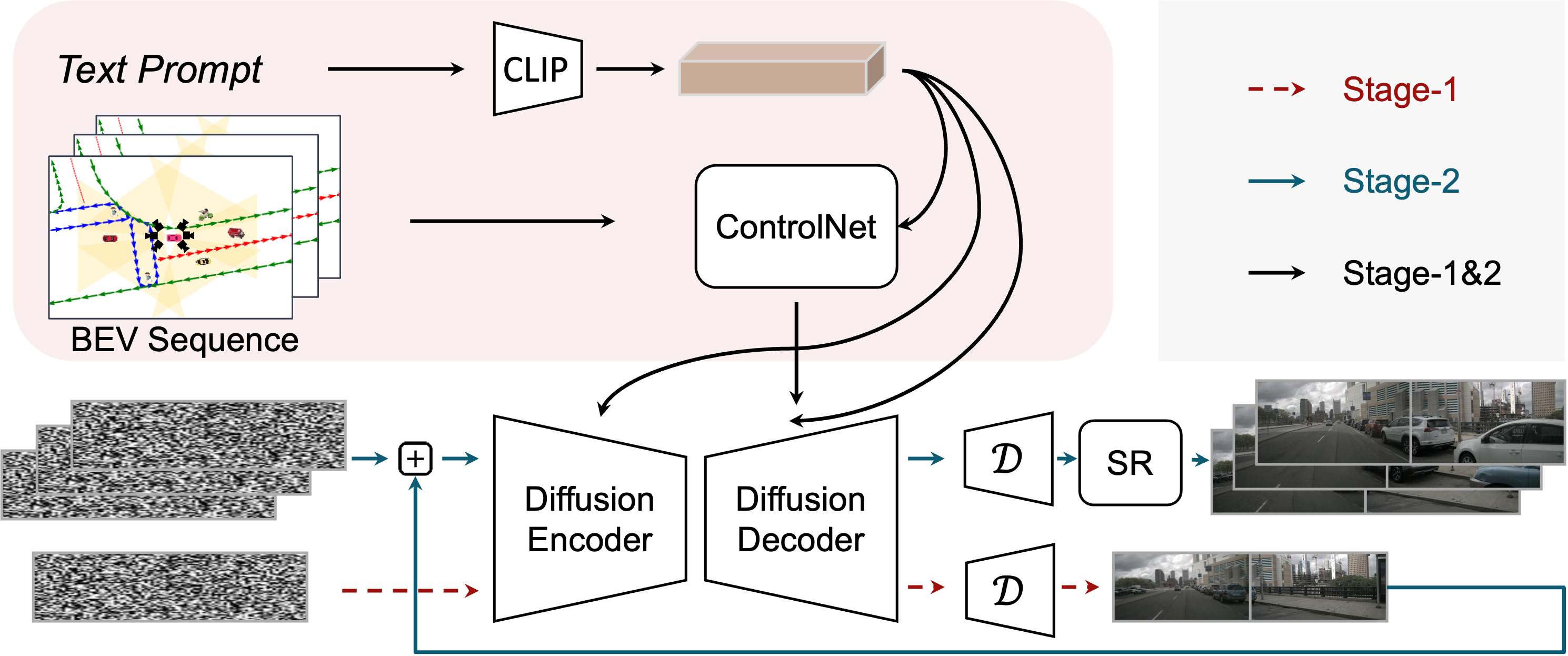}
   \hfill
    \caption{\textbf{The two-stage inference pipeline of Panacea+.} Its two-stage process begins by creating multi-view images with BEV layouts, followed by using these images, along with subsequent BEV layouts, to facilitate the generation of following frames. A super-resolution module is further appended to increase the resolution.
    }
   \label{fig:inference}
\end{figure}

During inference, as shown in Fig.~\ref{fig:inference}, we first sample multi-view frames using the weights of the first stage. This is followed by the generation of a multi-view video, which is conditioned on the initially generated frames, employing the weights of the second stage. This two-stage pipeline significantly enhances visual fidelity, a result attributable to the decomposition of spatial and temporal synthesis processes. 

\subsubsection{Multi-view Appearance Noise Prior}
To comprehensively enhance both temporal and view consistency, we adopt the method of multi-view appearance noise prior, following Microcinema~\cite{DBLP:journals/corr/abs-2311-18829}. Given that Panacea+ is a two-stage method, as presented in Sec.~\ref{two stage}, our framework allows the first frame of the video as a conditional input to generate subsequent frames. To ensure better content and appearance consistency between the subsequent frames and the conditional frame, as well as to improve consistency across different views, we integrate the features of the multi-view first frame images into the joint noise of the subsequent frames. Let $[x^1,x^2,\dots,x^N]$ represent the noise of multi-view frames with a 
 number of N within a video, where $x^i$ is the joint noise added to the i-th frame across multiple views, and $x^i_t$ is randomly sampled from a Gaussian distribution $\mathcal{N}(\mathbf{0}, \boldsymbol{I})$. If
$z^1$ is the latent feature of the first frame, then the training noise is:
\begin{equation}
\label{eq:noise prior}
x^i=\lambda z^1+x^i_t
\end{equation}
where $\lambda$ is the coefficient that controls the amount of the first multi-view frame features.

\subsection{Generating Controllable Driving Scene Videos} \label{control}

In our proposed Panacea+ model, designed for the advancement of autonomous driving systems, the controllability of synthesis samples emerge as a pivotal attribute. Panacea+ integrates two categories of control signals: a coarse-grained global control, encompassing textual attributes, and a fine-grained layout control, which involves BEV layout sequence. 

The coarse-grained global control endows the Panacea+ model to generate diverse multi-view videos.  This is achieved by integrating CLIP-encoded~\cite{DBLP:conf/icml/RadfordKHRGASAM21} text prompts into the UNet,  a method analogous to that used in Stable Diffusion. Benefiting from the Stable Diffusion pre-trained model, Panacea+ synthesizes specific driving scenes in response to textual prompts.

Panacea+'s fine-grained layout control facilitates the generation of synthesis samples that align with annotations. We use BEV layout sequences as the condition. Specifically, for a BEV sequence of duration T, we convert them into a perspective view and extract the control elements as object bounding boxes, object depth maps, road maps, and camera pose embeddings. Fig.~\ref{fig:train} (d) illustrates this process, where we employ different channels, represented by distinct colors, to delineate these segmented elements. This results in layout-controlling images with 19 channels: 10 for depth, 3 for bounding boxes, 3 for road maps, and 3 for camera pose embeddings. These 19-channel images are then integrated into the UNet using the ControlNet \cite{DBLP:journals/corr/abs-2302-05543} framework.

It is noteworthy that the camera pose essentially represents the direction vector~\cite{DBLP:conf/cvpr/ZhouK22,DBLP:conf/cvpr/00050GAG23,DBLP:journals/corr/abs-2301-04634}, which is derived from the camera's intrinsic and extrinsic parameters. 
Specifically, for each point $p^{c} = (u,v,d, 1)$ in camera frustum space, it can be projected into 3D space by intrinsic $K\in \mathbb{R}^{4 \times 4}$ and extrinsic $E\in \mathbb{R}^{4 \times 4}$ as in Eq.~\ref{eq_proj}:

\begin{equation}\label{eq_proj}
p^{3d} = E \times K^{-1} \times p^{c}
\end{equation}

Here $(u, v)$ denotes the pixel coordinates in the image, $d$ is the depth along the axis orthogonal to the image plane, $1$ is for the convenience of calculation in homogeneous form. We pick two points and set the depth of them to $d_1=1, d_2 =2$. Then the direction vector of the corresponding camera ray can be computed by:
\begin{equation}
\text{DV}(u,v) = p^{3d} (d_2) - p^{3d} (d_1)
\end{equation}
We normalize the direction vector by dividing the vector magnitude and multiply $255$ to simply convert it into RGB pseudo-color image.

\section{Experiment}

\subsection{Datasets and Evaluation Metrics} \label{controllability}

We evaluate the generation quality and controllability of Panacea+ and its benefits to autonomous driving on the nuScenes~\cite{DBLP:conf/cvpr/CaesarBLVLXKPBB20} dataset and the Argoverse 2~\cite{DBLP:journals/corr/abs-2301-00493} dataset. 

\noindent\textbf{nuScenes Dataset.} 
The nuScenes dataset is a public driving dataset that comprises 1000 scenes from Boston and Singapore. Each scene is a 20 seconds video with about 40 frames. It offers 700 training scenes,  150 validation scenes, and 150 testing scenes with 6 camera views. The camera views overlap each other, covering the whole 360-degree field of view.

\noindent\textbf{Argoverse 2 Dataset.} 
The Argoverse 2 dataset is a public  dataset for autonomous driving, containing 1000 scenes, each is of 15 seconds in length, with annotations at a frequency of 10Hz. It is divided into 700 scenes for training, 150 for validation, and 150 for testing. It is sourced from 7 high-resolution cameras that cover a 360-degree field of view. We evaluate it with 10 categories.

\noindent\textbf{Generation Quality Metrics.}
We utilize the frame-wise Fréchet Inception Distance (FID)~\cite{DBLP:conf/nips/HeuselRUNH17} and the Fréchet Video Distance (FVD)~\cite{DBLP:journals/corr/abs-1812-01717} to evaluate the quality of our synthesised data, where FID reflects the image quality and FVD is a temporal-aware metric that reflects both the image quality and temporal consistency.

\noindent\textbf{Controllability Metrics.} 
The controllability of Panacea+ is demonstrated through the alignment between the generated videos and the conditioned BEV sequences. To substantiate this alignment, we assess the perceptual performance on the nuScenes and Argoverse 2 datasets. For the 3D object tracking task, we employ metrics such as Average Multi-Object Tracking Accuracy (AMOTA), Average Multi-Object Tracking Precision (AMOTP), and Multi-Object Tracking Accuracy (MOTA). For 3D object detection, we utilize the nuScenes Detection Score (NDS), mean Average Precision (mAP), mean Average Orientation Error (mAOE), and mean Average Velocity Error (mAVE) on the nuScenes dataset, and the Composite Detection Score (CDS) and mAP on the Argoverse 2 dataset. StreamPETR, a state-of-the-art video-based perception method, serves as our main evaluation tool for nuScenes. For Argoverse 2, we use Far3D~\cite{DBLP:conf/aaai/JiangLLWJWHZ24F}  as our evaluation model. For lane detection, we assess mAP using MapTR~\cite{DBLP:conf/iclr/LiaoCWC00H23}. To evaluate the controllability of our generated samples, we compare the validation performance of our generated data against real data using the pre-trained perception models. The potential to augment the training set as a strategy for performance enhancement further reflects the controllability of Panacea+.

\begin{table}[tb]
    \centering
    \footnotesize
        \centering
         \caption{Comparing FID and FVD metrics with SoTA methods on the validation set of the nuScenes dataset. }
        \label{tab:fvd}
        \resizebox{0.475\textwidth}{!}{
        \setlength{\tabcolsep}{8pt}
        \begin{tabular}{lcccc}
            \toprule
            Method     &Multi-View &Multi-Frame  & FVD$\downarrow$ & FID$\downarrow$ \\
            \hline
            BEVGen\cite{DBLP:journals/corr/abs-2301-04634} \  &$\checkmark$   &  &  & 25.54\\
            BEVControl\cite{DBLP:journals/corr/abs-2308-01661}  &$\checkmark$   &  & - & 24.85 \\
            MagicDrive\cite{DBLP:journals/corr/abs-2310-02601}  &$\checkmark$   &  & - & 16.20\\
            WoVoGen\cite{DBLP:journals/corr/abs-2312-02934}   &$\checkmark$ &$\checkmark$  & 417.7 & 27.6\\
            DriveDreamer\cite{DBLP:journals/corr/abs-2309-09777} \footnotemark[1]  &$\checkmark$  &$\checkmark$  & 353 & 26.8\\
            Panacea\cite{DBLP:journals/corr/abs-2311-16813}   &$\checkmark$ &$\checkmark$  & 139 & 16.96\\
            SubjectDrive\cite{DBLP:journals/corr/abs-2403-19438}   &$\checkmark$ &$\checkmark$  & 124 & 15.96\\
            DriveDreamer-2\cite{DBLP:journals/corr/abs-2403-06845}   &$\checkmark$ &$\checkmark$  & 105 & 25.0\\
            \hline
             \rowcolor[gray]{.9}Panacea+   &$\checkmark$  &$\checkmark$  & 103 & 15.5  \\
            \bottomrule
        \end{tabular}
        }

\end{table}

\footnotetext[1]{Arxiv v2 version.}
\subsection{Implementation Details}

We implement our approach based on Stable Diffusion 2.1~\cite{DBLP:conf/cvpr/RombachBLEO22} . Pre-trained weights are used to initialize the spatial layers in UNet. During our two-stage training, the image weights of the first stage are optimized for 56k steps, and the video weights of the second stage is optimized for 40k steps. For inference, we utilize a DDIM~\cite{DBLP:conf/iclr/SongME21} sampler configured with 25 sampling steps. The video samples are generated at a spatial resolution of $256\times512$ by the diffusion model, with a frame length of 8. Then the super-resolution module upscales the frames spatially to $512\times1024$. The appearance noise prior coefficient is 0.05 for training and 0.07 during the inference stage. 
For evaluation, we use StreamPETR for 3D object detection and tracking on nuScenes, building upon a ResNet50~\cite{DBLP:conf/cvpr/HeZRS16} backbone. We use Far3D on Argoverse 2 for object detection, with ResNet50 as backbone. For lane detection, we use MapTR with R18~\cite{DBLP:conf/cvpr/HeZRS16} as backbone.

\subsection{Quantitative Analysis}
\subsubsection{Generation Quality} 
We initially employ the quantitative metrics FVD and FID to assess the quality and fidelity of the generated multi-perspective videos, comparing them with SOTA methods in the autonomous driving domain for image and video synthesis. As shown in Tab.~\ref{tab:fvd}, our approach achieves the best FVD and FID scores compared to all state-of-the-art (SOTA) methods. When compared with image synthesis methods, even though our task is more challenging video generation, our FID still holds an advantage. For instance, we have a 0.7 point advantage over MagicDrive. Compared to BEVGen and BEVControl, we also exhibit a significant advantage. Furthermore, when compared with video synthesis methods, our method also demonstrates superior performance. Our approach is dedicated to synthesizing video samples with excellent temporal consistency, and thus, FVD can effectively reflect the quality of our sample consistency. It can be observed that our FVD improves by 36 points compared to our previous work, Panacea, indicating an improvement in temporal consistency. When compared with other SOTA methods, our approach also perform well. For example, our FVD score is 21 points lower than SubjectDrive. In terms of FID compared to these video synthesis methods, we also hold an advantage. For instance, we achieve slightly better performance than Drivedreamer-2 in FVD, while holding a  significantly superior improvement in FID, indicating the high quality of our videos.

\begin{table}[tb]
    \centering
    \footnotesize
        \centering
         \caption{Comparison of the generated data with real data on the validation set, employing a pre-trained perception model.}
        \resizebox{0.32\textwidth}{!}{
        \setlength{\tabcolsep}{7pt}
        \begin{tabular}{lcc}
            \toprule
              Method  &Resolution  &  NDS$\uparrow$\\
            \hline
            \hline
             Real &512$\times$256  &46.9 \\ 
             Panacea&512$\times$256   &32.1(68\%)  \\
               \rowcolor[gray]{.9} Panacea+ & 512$\times$256  &  34.6 (74\%) \\
            \bottomrule
        \end{tabular}
        }
        \label{tab:valset}
\end{table}

\subsubsection{Controllability}
We utilize the pre-trained StreamPETR to evaluate the controllability of the generated samples. We test the detection metric NDS on both the real and synthetic nuScenes validation sets. By comparing their relative performance, we measure the alignment of our generated samples with BEV layouts. This degree of alignment reflects the controllability of our video synthesis.
As can be seen from Table \ref{tab:valset}, the validation set synthesized by Panacea+ achieved an NDS of 34.6, which corresponds to 74\% of the NDS achieved by the real dataset. This indicates that a significant portion of our generated video samples is aligned with the BEV control signals. Moreover, compared to Panacea, we have a gain of 2.5 points. This demonstrates an improvement in the controllability of the generated videos, which brings benefits in terms of alignment. Therefore, our synthetic data samples have the potential to provide greater assistance to downstream tasks, especially those strongly correlated with temporal dynamics, like tracking in autonomous driving.

\begin{table}[tb]
    \centering
    \footnotesize
        \centering
         \caption{Comparison involving data augmentation using synthetic data on 3D object tracking task.}
        \resizebox{0.475\textwidth}{!}{
        \setlength{\tabcolsep}{4pt}
        \begin{tabular}{lcccccc}
            \toprule
            Method&Resolution&AMOTA$\uparrow$ & MOTA$\uparrow$ & AMOTP$\downarrow$ \\
            \hline
            Real only&512$\times$256&30.1	& 27.1  & 1.379 \\
       
           Real+Panacea&512$\times$256& 33.7 &30.6  &1.353\\
            Real+Panacea+&512$\times$256&34.6&31.1& 1.343 \\
            
            \hline
            Real only&1024$\times$512 & 37.6	& 32.0 & 1.285\\
            \rowcolor[gray]{.9}  Real+Panacea+& 1024$\times$512& 42.7 &37.7 & 1.233  \\
            \bottomrule
        \end{tabular}
        }
        \label{tab:augment-track}
\end{table}

\begin{table}[tb]
    \centering
    \footnotesize
        \centering
         \caption{Comparison involving data augmentation using synthetic data on 3D object detection task.}
        \resizebox{0.475\textwidth}{!}{
        \setlength{\tabcolsep}{4pt}
        \begin{tabular}{lccccccc}
            \toprule
            Method&Resolution&mAP$\uparrow$ & mAOE$\downarrow$ & mAVE$\downarrow$ & NDS$\uparrow$ \\
            \hline
            Real only&512$\times$256& 34.5	& 59.4 & 29.1  & 46.9 \\
          
           Real+Panacea&512$\times$256& 37.1 &54.2 &27.3 &49.2\\

           Real+Panacea+& 512$\times$256& 37.1 &51.0 & 28.1 &49.4 \\
            \hline
            Real only&1024$\times$512& 40.4	& 49.4 & 30.0  & 51.2 \\
            \rowcolor[gray]{.9} Real+Panacea+& 1024$\times$512 & 42.6 &40.9 & 27.1 &53.8\\
            \bottomrule
        \end{tabular}
        }
        \label{tab:augment-detect}
\end{table}

\begin{table}[tb]
    \centering
    \footnotesize
        \centering
        \caption{Comparison with SoTA methods on 3d object detection and tracking tasks.}
        \resizebox{0.4\textwidth}{!}{
        \setlength{\tabcolsep}{4pt}
        \begin{tabular}{lcc|cc}
            \toprule
            Method &mAP$\uparrow$ & NDS$\uparrow$ &AMOTA$\uparrow$ & AMOTP$\downarrow$ \\
            \hline
           
            DriveDreamer-2\cite{DBLP:journals/corr/abs-2403-06845}& 32.9 &45.4 &31.3&1.387\\
            MagicDrive\cite{DBLP:journals/corr/abs-2310-02601}& 35.4 &39.8 &- &-\\
            DriveDreamer\cite{DBLP:journals/corr/abs-2309-09777}& 35.8 &39.5 &- &-\\
            WoVoGen\cite{DBLP:journals/corr/abs-2312-02934}& 36.2 &18.1 &- &-\\
           Panacea\cite{DBLP:journals/corr/abs-2311-16813}& 37.1 &49.2 &33.7 &1.353\\
           SubjectDrive\cite{DBLP:journals/corr/abs-2403-19438}& 38.1 &50.2 &37.2 &1.317\\
           \rowcolor[gray]{.9}Panacea+ & 42.6  &53.8 &42.7 &1.233 \\
            \bottomrule
        \end{tabular}
        }
        \label{augment-sota}
\end{table}

\subsubsection{Enhancement for Autonomous Driving Tasks} ~\\
Given that our ultimate goal is to aid autonomous driving systems, we use our generated samples as training resources for perception models. To ensure the completeness of our validation, we conducted experiments across multiple downstream tasks, including 3D object tracking, 3D object detection, and lane detection. Among these tasks, tracking best reflects the importance of temporal consistency.
All our experiments use a baseline model trained exclusively with real datasets. We then introduce an equal amount of synthetic data into the training process to evaluate the benefits it provides. \\
\noindent \textbf{Performance on 3D Object Tracking Task.}
As depicted in Table \ref{tab:augment-track}, at a resolution of 512 $\times$ 256, training with the generated samples from Panacea+ yields an AMOTA of 34.6, which is 3.5 points higher than the baseline trained solely with real data, representing a substantial performance improvement. Additionally, a 0.9 point gain compared to Panacea indicates that Panacea+ indeed enhances consistency.
Furthermore, we explore the impact of higher resolutions and observe that at 1024 $\times$ 512 resolution, the absolute performance peakes at 42.7, which is 5.1 points higher than the baseline at the same resolution and 12.6 points higher than the baseline at 512 $\times$ 256 resolution. This not only shows that synthetic samples with larger resolutions more effectively assist tracking tasks, but also proves that our synthetic data achieves excellent relative performance on a higher baseline. 
When compared with state-of-the-art methods, it can be observed from Tab.~\ref{augment-sota} that, Panacea+ achieves the best performance in tracking tasks compared to previous methods as a holistic model for synthesizing high-resolution, high-quality video samples. For instance, it outperforms DriveDreamer-2 by 11.4 points in AMOTA. This significant margin underscores Panacea+'s superior capability to generate high-fidelity video samples that not only excel in quality but also significantly advance tracking capabilities in autonomous driving.

\noindent \textbf{Performance on 3D Object Detection Task.}
We also conduct experiments on 3D object detection task. When trained solely on the generated video samples, the StreamPETR model achieved a notable NDS of 38.4, which is 82\% of the performance relative to the baseline trained on the real nuScenes training set. As shown in Tab.~\ref{tab:augment-detect},  the fusion of synthetic and real data elevates the StreamPETR model to an NDS of 49.4, outperforming the model trained exclusively on real data by a margin of 2.5 points. Moreover, since our overall framework can generate samples with larger resolutions, we explore the gains brought by further increasing the resolution. We conduct experiments at a resolution of 1024 $\times$ 512 and find that a larger resolution leads to significant performance improvements, achieving an NDS of 53.8. This not only validates the effectiveness of our samples under different baseline settings but also demonstrates their strong potential at high resolutions. In Tab.~\ref{augment-sota}, the comparison with SOTA methods also showcases the exceptional performance of Panacea+. We achieve an MAP of 53.8, surpassing all current autonomous driving scene generation methods. This achievement highlights the effectiveness of Panacea+. For the Argoverse 2 dataset, we also conduct evaluation on the augmented dataset training for 3D object detection. It can be observed from Tab.~\ref{tab:all} that the CDS and mAP increase by 1.1 and 0.9 points, respectively. This indicates that our generation model is also effective across different datasets.

\begin{table}[tb]
    \centering
    \footnotesize
        \centering
                 \caption{Comparison involving data augmentation using synthetic data on multiple tasks and datasets.}
        \resizebox{0.45\textwidth}{!}{
        \setlength{\tabcolsep}{2.0pt}
                  \begin{tabular}{c|c|cc|cc}
    \toprule
      Task& Dataset&Real & Generated&\multicolumn{2}{c}{Metrics}\\
              \midrule
    \multirow{3}*{Object Detection}&\multirow{3}*{Argoverse v2}&\multicolumn{2}{c|}{}&CDS$\uparrow$ & mAP$\uparrow$\\
    \cline{5-6}
     &&$\checkmark$ & - & 17.1& 23.6 \\
     &&$\checkmark$ &  $\checkmark$ & 18.2 \textcolor{blue}{(+1.1)} & 24.7 \textcolor{blue}{(+0.9)}\\
     \hline
   
     \multirow{3}*{Object Detection}&\multirow{3}*{nuScenes}&\multicolumn{2}{c|}{}&NDS$\uparrow$& mAP$\uparrow$\\
    \cline{5-6}
     &&$\checkmark$ & - & 51.2& 40.4\\
     &&$\checkmark$ &  $\checkmark$ & 53.8 \textcolor{blue}{(+2.6)} & 42.6 \textcolor{blue}{(+2.2)}\\
          \hline
     \multirow{3}*{Object Tracking}&\multirow{3}*{nuScenes}&\multicolumn{2}{c|}{}&AMOTA$\uparrow$ & AMOTP$\downarrow$\\
    \cline{5-6}
     &&$\checkmark$ & - & 37.6& 1.285\\
     &&$\checkmark$ &  $\checkmark$ & 42.7 \textcolor{blue}{(+5.1)} & 1.233 \textcolor{blue}{(-0.052)}\\
               \hline

     \multirow{3}*{Lane Detection}&\multirow{3}*{nuScenes}&\multicolumn{2}{c|}{}&-&mAP$\uparrow$\\
    \cline{5-6}
     &&$\checkmark$ & - & -&45.0\\
     &&$\checkmark$ &  $\checkmark$ & -&51.0 \textcolor{blue}{(+6.0)}\\
    \bottomrule
  \end{tabular}
        }
        \label{tab:all}
\end{table}

\noindent \textbf{Performance on Lane Detection Task.}
By integrating road map control signals into our synthetic samples, we significantly enhance the performance of lane detection task. We leverage the synthetic data to train the advanced perception model MapTR on the nuScenes dataset. As illustrated in Tab.\ref{tab:all}, we achieve a notable improvement of 6.0 points. This substantial enhancement highlights the exceptional capabilities of our model in lane detection, demonstrating its potential to markedly improve the performance of existing perception models.

\begin{figure}
  \centering
   \includegraphics[width=1.0\linewidth]{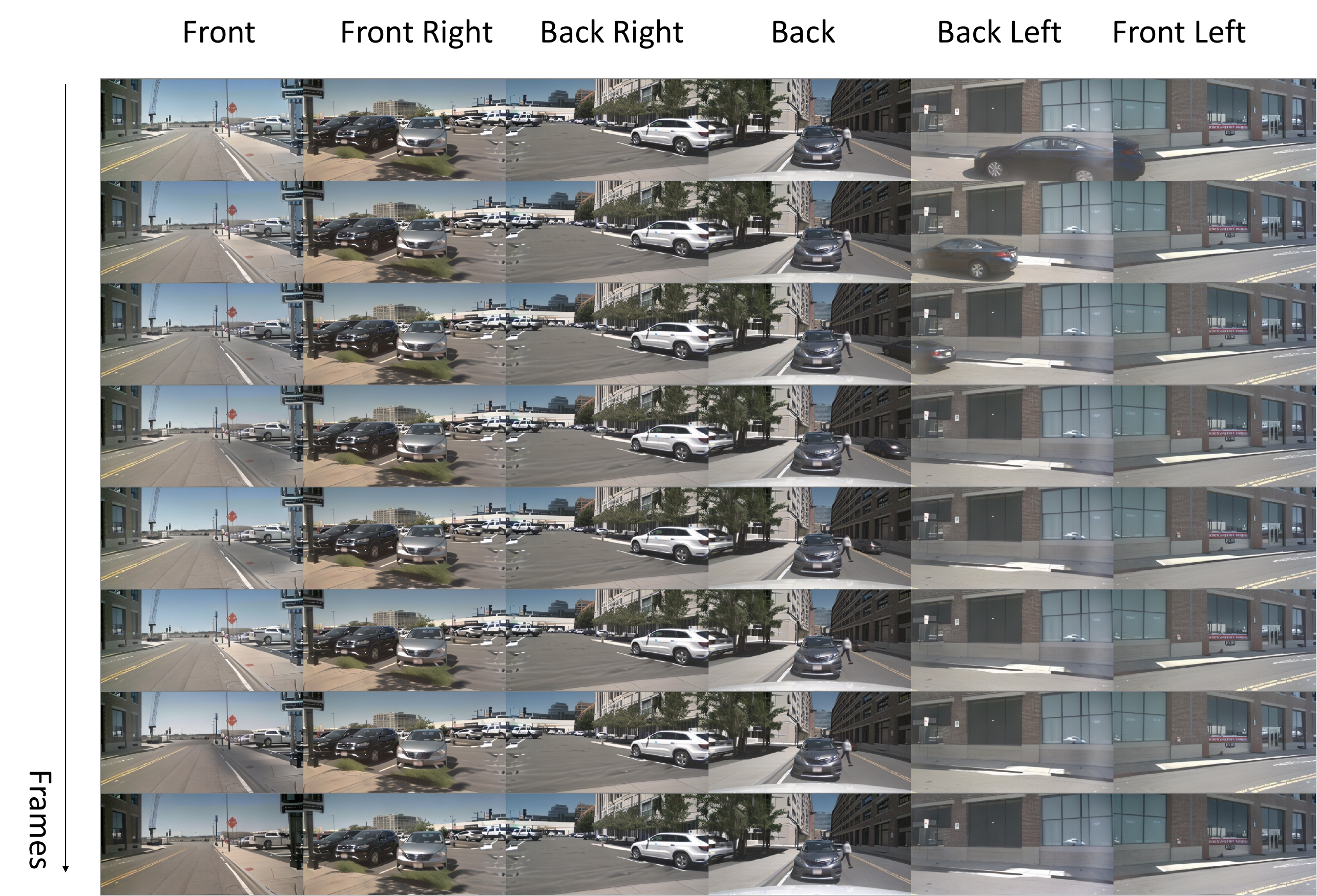}
   \hfill
   \caption{Generated videos generated by Panacea+ on nuScenes dataset.}
   \label{nus}
   \vspace{-0.5cm}
\end{figure}

\subsection{Qualitative Analysis }
\noindent \textbf{Generation Quality.} 
As depicted in Fig.~\ref{nus}. Panacea+ demonstrates the ability to generate realistic multi-view videos from BEV sequences and text prompts. The generated videos exhibit notable temporal and cross-view consistency. We present all views across eight subsequent frames to fully showcase the video samples. It can be seen that Panacea+ successfully synthesizes high-quality samples, maintaining both temporal and view consistency.

\begin{figure}
  \centering
   \includegraphics[width=\linewidth]{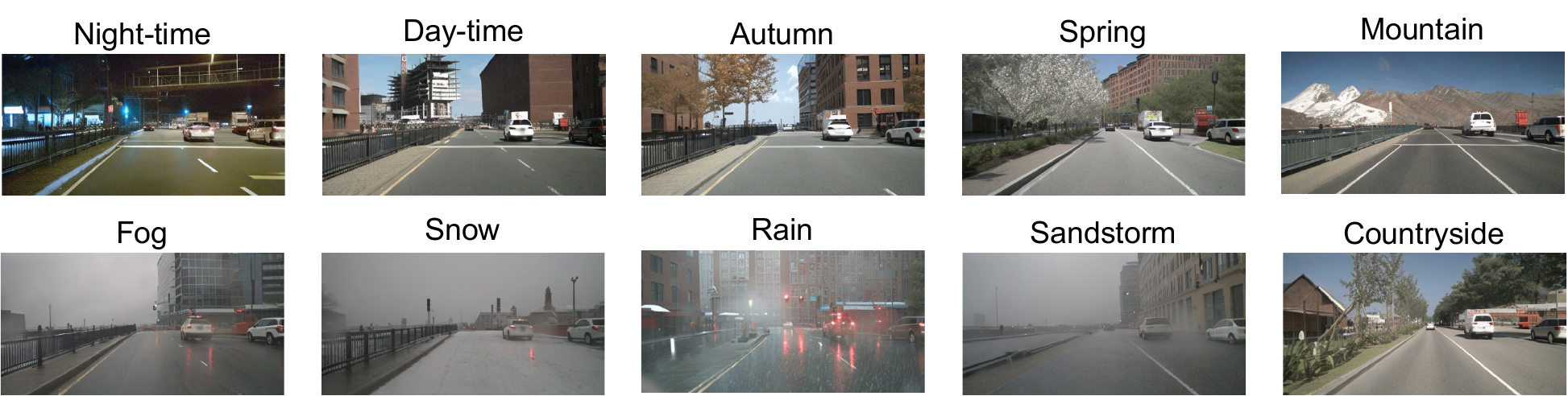}
   \hfill
   \caption{Controllable generation with textual attributes.}
   \label{demo weather}
\end{figure}

\begin{figure}
  \centering
   \includegraphics[width=\linewidth]{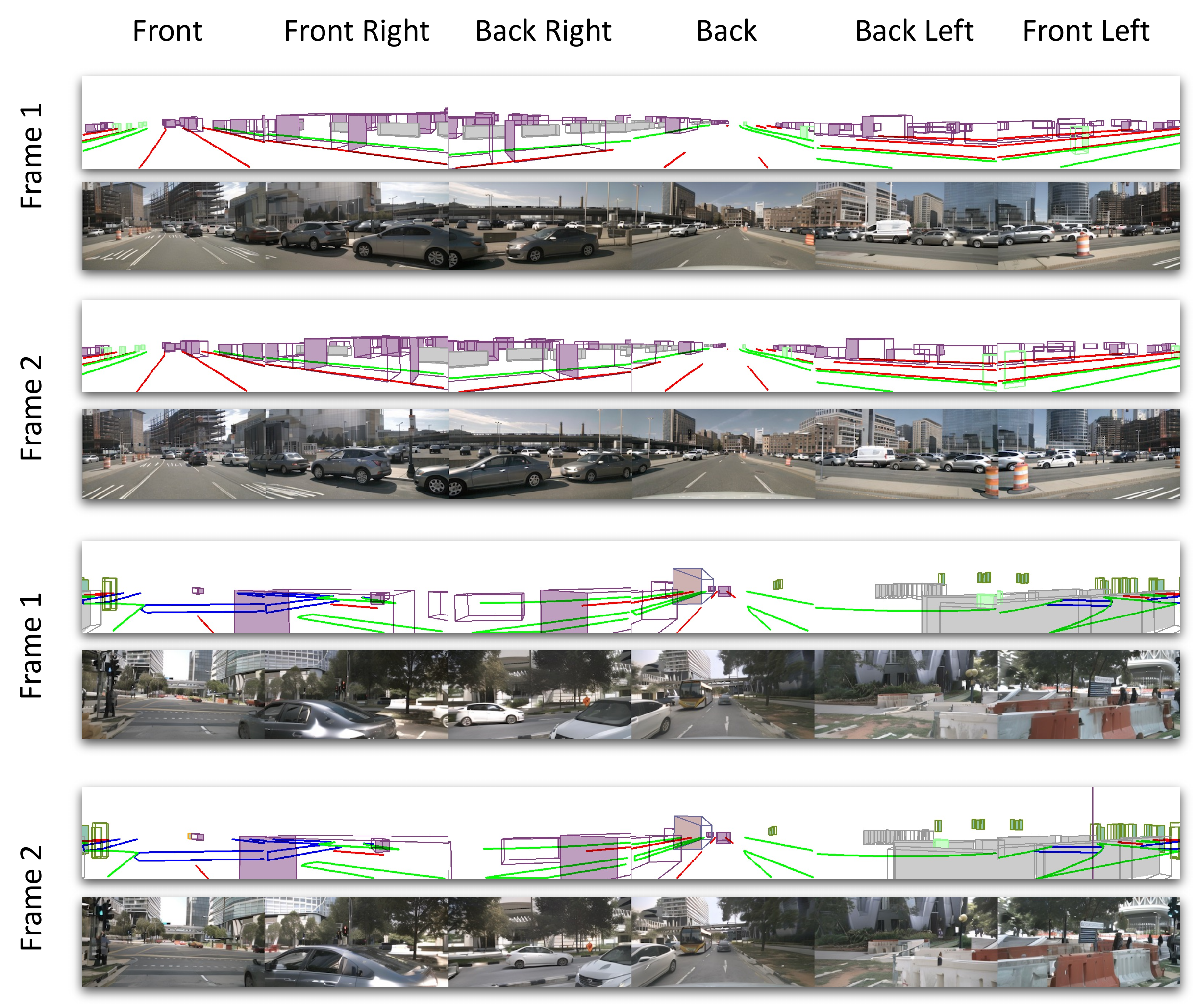}
   \hfill
   \caption{Controllable multi-view video generation with BEV layouts. The column shows 6 different views and the rows shows adjacent frames aligned with BEV control.}
   \label{demo align}
\end{figure}

\noindent \textbf{Controllability for Autonomous Driving.} We visualize the controllability of Panacea+ from two aspects. The first involves coarse-grained textual control. Fig.~\ref{demo weather} illustrates the attribute control capabilities, showing how modifications to text prompts can manipulate elements such as weather, time, and scene. This flexibility enables our approach to simulate a variety of rare driving scenarios, including extreme weather conditions like rain and snow, thereby significantly enhancing the diversity of the data.
Additionally, Fig.~\ref{demo align} depicts how cars and roads align precisely with the BEV layouts while maintaining excellent temporal and view consistency, showcasing the fine-grained controllability of Panacea+.

\subsection{Ablations. }
\subsubsection{Ablations on Multi-view Appearance Noise Prior}
This section validates the effectiveness of the multi-view appearance noise prior and conducts ablation studies on its strength to verify our experimental settings. We compare FVD and NDS scores for generation quality and controllability respectively to choose a $\lambda$ value that offers the best overall performance. When $\lambda$ equals zero, it indicates that no appearance noise prior is added. We observed that by adding more noise prior in a suitable range during inference, the generation quality and controllability can be enhanced simultaneously. Compared to setups without the noise prior, the performances are better. We choose 0.07 as the optimal $\lambda$ value for generation.

\begin{table}[tb]
    \centering
    \caption{Ablation study on the choice of $\lambda$}
    \footnotesize
        \centering
        \resizebox{0.13\textwidth}{!}{
        \setlength{\tabcolsep}{2.0pt}
                  \begin{tabular}{c|cc}
    \toprule
     $\lambda$ & FVD & NDS\\
    \midrule
    0 &139 & 32.1 \\
     0.05 & 125 & 33.8\\
    0.06 &  99 & 34.3\\
     \rowcolor[gray]{.9} 0.07 & 103 & 34.6\\
    0.08 & 105 & 33.6\\
    \bottomrule
  \end{tabular}
        }
        \label{tab:expand-map}
\end{table}

\subsubsection{Ablations on Super-Resolution Module}
We have considered two super-resolution methods: one using the stableSR~\cite{DBLP:journals/corr/abs-2305-07015} network for 2$\times$ spatial up-sampling, and the other a simple resize. We use both methods for the expansion of training in tracking and detection tasks. It can be observed that in tracking tasks, the network using StableSR holds a 0.8 point gain in AMOTA compared with resize, while in detection tasks, the performance of both methods is similar. Therefore, in practical use, if considering computational cost, the simple resize method is actually a more cost-effective choice. Using a super-resolution module incurs additional computational costs, but it can also yield certain performance improvements.
\begin{table}[tb]
    \centering
    \footnotesize
        \centering
         \caption{Ablation study on the super-resolution module.}
        \resizebox{0.475\textwidth}{!}{
        \setlength{\tabcolsep}{4pt}
        \begin{tabular}{lccccccccc}
            \toprule
            Method&  AMOTA$\uparrow$ & MOTA$\uparrow$ & AMOTP$\downarrow$  &mAP$\uparrow$ & mAOE$\downarrow$ & mAVE$\downarrow$ & NDS$\uparrow$ \\
            \hline
           Real Only&37.6	& 32.0 & 1.285 & 40.4	& 49.4 & 30.0  & 51.2 \\
            Real+Panancea+ (Resize)& 41.9 &36.3 & 1.230 &42.6 &40.2 &28.2 & 53.7\\
            \rowcolor[gray]{.9}  Real+Panancea+  (StableSR) & 42.7 &37.7 & 1.233  & 42.6 &40.9 & 27.1 &53.8\\
            \bottomrule
        \end{tabular}
        }
        \label{tab:ablation-high}
\end{table}

\section{Conclusion}
We propose Panacea+, a powerful and versatile data generation framework to create controllable panoramic videos for driving scenarios. Within this framework, we incorporate a decomposed 4D attention module to ensure temporal and cross-view consistency and adopt multi-view appearance noise prior to further enhance consistency. Additionally, a two-stage strategy is employed to improve the generation quality. Panacea+ is adept at handling a variety of control signals to produce videos with precise annotations. Moreover, Panacea+ also integrated a super-resolution module to explore the benefits to perception models at high resolution.  Through extensive experiments across various tasks and datasets, Panacea+ has demonstrated its proficiency in generating valuable videos, serving a wide spectrum of BEV perception.

\textbf{Limitations:} Panacea+ still offers considerable room for exploration. Currently, our experiments rely on videos synthesized from the labeled signals of existing datasets. In the future, we could integrate with simulators or develop methods for generating control signals. Additionally, due to our relatively high computational costs, future explorations could focus on adopting more computationally efficient methods. Moreover, there is potential for further study in scaling the model, especially for diffusion models using the transformer structure, which currently represents a promising direction.

{\small
\bibliographystyle{IEEEtran}
\bibliography{egbib}
}

\vspace{10pt}

\noindent \textbf{Yuqing Wen} received the B.E. degree from  University of Science and Technology
of China, Hefei, Anhui, China. She is currently working toward the Ph.D. degree with University of Science and Technology
of China, Hefei, Anhui, China.

\vspace{10pt}

\noindent \textbf{Yucheng Zhao} received B.E. degree and the Ph.D. degree with University of Science and Technology
of China, Hefei, Anhui, China. He is currently a researcher with MEGVII Technology, Beijing, China.

\vspace{10pt}

\noindent \textbf{Yingfei Liu} Yingfei Liu received the B.Sc. degree from Tianjin University, Tianjin, China. He received the master’s degree with the Aerospace Information Research Institute, Chinese Academy of Sciences, Beijing, China. He is currently a researcher with MEGVII Technology, Beijing, China.

\vspace{10pt}

\noindent \textbf{Binyuan Huang} Binyuan Huang received the B.Eng. degree from South China Normal University, China, in 2023. He is currently a Master at the Wuhan University, China.

\vspace{10pt}

\noindent \textbf{Fan Jia} recieved the master’s degree from Tianjin University, Tianjin, China. He is currently a researcher with MEGVII Technology, Beijing, China.

\vspace{10pt}

\noindent \textbf{Yanhui Wang} received the B.E. degree from  University of Science and Technology of China, Hefei, Anhui, China. He is currently working toward the Ph.D. degree with University of Science and Technology
of China, Hefei, Anhui, China.

\vspace{10pt}

\noindent \textbf{Chi Zhang} obtained Master degree from Tsinghua University. He received PHD degree from Columbia University. He is currently a researcher at Mach Drive, Wuhu, Anhui, China.

\vspace{10pt}

\noindent \textbf{Tiancai Wang} obtained Master degree from Tianjin University. He was a research intern at MEGVII Research, Beijing, China and Inception Institute of Artificial Intelligence (IIAI), Abu Dhabi, UAE. He is currently a researcher at MEGVII Technology, Beijing, China.

\vspace{10pt}

\noindent \textbf{Xiaoyan Sun} (Senior Member, IEEE) received the B.S., M.S., and Ph.D. degrees in computer science from the Harbin Institute of Technology, Harbin, China, in 1997, 1999, and 2003, respectively.She is currently a Full Professor with University of Science and Technology of China, Hefei, China, and the Deputy Director of National Engineering Laboratory for Brain-inspired Intelligence Technology and Application. Dr. Sun was a recipient of the Best Paper Award of IEEE Transactions on Circuits and Systems for Video Technology in 2009 and the Best Student Paper Award of VCIP 2016. She is on the Senior Editorial Board of IEEE Journal on Emerging and Selected Topics in Circuits and Systems and AE of Signal Processing: Image Communication.

\vspace{10pt}

\noindent \textbf{Xiangyu Zhang} received the PhD degree from
Xi’an Jiaotong University, in 2017. He is currently
a senior researcher with MEGVII Research, Beijing, China. He got
CVPR Best Paper Award, in 2016. He also serves
as a reviewer for CVPR, ICCV, ECCV, NeurIPS,
ICLR, the IEEE Transactions on Pattern Analysis
and Machine Intelligence, etc. The total number
of his Google Scholar citations is more than
290,000.

\end{document}